\documentclass{article}

\usepackage{amsmath,amsfonts,amssymb,times,graphicx,natbib,algorithm,algorithmic,hyperref}
\usepackage{booktabs} 
\usepackage{multirow}
\usepackage[caption=false, font=footnotesize]{subfig}

\usepackage[accepted]{data4good2016}
\usepackage{tikz}
\icmltitlerunning{Improving Power Generation Efficiency using Deep Neural Networks}

\begin{document}

\twocolumn[
\icmltitle{Improving Power Generation Efficiency using Deep Neural Networks}

\icmlauthor{Stefan Hosein}{stefan.hosein2@my.uwi.edu}
\icmlauthor{Patrick Hosein}{patrick.hosein@sta.uwi.edu}
\icmladdress{The University of the West Indies, St. Augustine, Trinidad and Tobago}

\vskip 0.3in
]

\begin{abstract}
Recently there has been significant research on power generation, distribution and transmission efficiency especially in the case of renewable resources. The main objective is reduction of energy losses and this requires improvements on data acquisition and analysis. In this paper we address these concerns by using consumers' electrical smart meter readings to estimate network loading and this information can then be used for better capacity planning. We compare Deep Neural Network (DNN) methods with traditional methods for load forecasting. Our results indicate that DNN methods outperform most traditional methods. This comes at the cost of additional computational complexity but this can be addressed with the use of cloud resources. We also illustrate how these results can be used to better support dynamic pricing.
\end{abstract}

\section{Introduction}

Currently, most of the energy produced worldwide uses coal or natural gas. However, much of this energy is wasted. In the United States of America, approximately 58\% of energy produced is wasted \cite{batt_2013}. Furthermore, 40\% of this wasted energy is due to industrial and residential buildings. By reducing energy wastage in the electric power industry, we reduce damage to the environment and reduce the dependence on fossil fuels. 

Short-term load forecasting (STLF) (i.e., one hour to a few weeks) can assist since, by predicting load,  one can do more precise planning, supply estimation and price determination. This leads to decreased operating costs, increased profits and a more reliable electricity supply for the customer. Over the past decades of research in STLF there have been numerous models proposed to solve this problem. These models have been classified into classical approaches like moving average \cite{Andrade_09} and regression models \cite{Hong_11}, as well as machine learning based techniques, regression trees \cite{Mori_01}, support vector machines \cite{Niu2006} and Artificial Neural Networks \cite{Lee_92}.



In recent years, many deep learning methods have been shown to achieve state-of-the-art performance in various areas such as speech recognition \cite{Hinton_12}, computer vision \cite{Krizhevsky_2012} and natural language processing \cite{Collobert_2008}. This promise has not been demonstrated in other areas of computer science due to a lack of thorough research. Deep learning methods are representation-learning methods with multiple levels of representation obtained by composing simple but non-linear modules that each transform the representation at one level (starting with the raw input) into a representation at a higher, slightly more abstract level \cite{LeCun_15}. With the composition of enough such transformations, very complex functions can be learned. 


In this paper, we compare deep learning and traditional methods when applied to our STLF problem and we also provide a comprehensive analysis of numerous deep learning models. We then show how these methods can be used to assist in the pricing of electricity which can lead to less energy wastage. To the best of our knowledge, there is little work in such comparisons for power usage in an electrical grid. The data we use is based on one year of smart meter data collected from residential customers. We apply each of the deep and traditional algorithms to the collected data while also noting the corresponding computational runtimes. Due to differences in electricity usage between the week and the weekend, we then split the data into two new datasets: weekends and weekly data. The algorithms are applied to these new datasets and the results are analyzed. The results show that the deep architectures are superior to the traditional methods by having the lowest error rate, but they do have the longest run-time. Due to space limitations we do not provide details of the traditional approaches but do provide references.

\section{Analysis}


\subsection{Data Description}
Our dataset consists of 8592 samples of 18 features that were collected from several households. The dataset was broken into 3 parts for training, validation and testing of sizes 65\%, 15\%, 20\% respectively. The readings were recorded at hourly intervals throughout the year. Some of the features were electrical load readings for the previous hour, the previous two hours, the previous three hours, the previous day same hour, the previous day previous hour, the previous day previous two hours, the previous 2 days same hour, the previous 2 days previous hour, the previous 2 days previous two hours, the previous week same hour, the average of the past 24 hours and the average of the past 7 days. The rest of the features (which do not contain electrical load readings) are the day of the week, hour of the day, if it is a weekend, if it is a holiday, temperature and humidity. These features were selected as they are typically used for STLF. In addition, the total electrical load does not change significantly throughout the year since the households are located in a tropical country where the temperature remains fairly constant throughout the year.

\begin{table}[!t]
\caption{Baseline algorithms}
\label{tab:1}
\centering
\begin{tabular}{rrrr}
\toprule
Algorithm & MAPE & MPE & Time (s)\\
\midrule
WMA 	&  9.51 & -1.96 & 100\\ 
MLR & 24.25	& -1.47 & 1\\ 
MQR & 12.91 & -7.63& 7\\
RT 	& \textbf{7.23}& -1.71& 15\\
SVR & 13.65	& \text{ }3.16 &19\\
\bottomrule
\end{tabular}
\end{table}

\subsection{Comparison Method}
As a preprocessing step, the data is cleaned and scaled to zero mean and unit variance. All traditional methods use cross-validation to determine appropriate values for the hyper-parameters. A random grid search was used to determine the hyper-parameters for the deep learning methods.

Several baseline algorithms were chosen. They include the Weighted Moving Average (WMA) where $y_{t+1} = \alpha y_{i} + \beta y_{i-167}$ with $\alpha = 0.05$ and $\beta = 0.95$, Multiple Linear Regression (MLR) and quadratic regression (MQR), Regression Tree (RT) with the minimum number of branch nodes being 8, Support Vector Regression (SVR) with a linear kernel and Multilayer Perception (MLP), with the number of hidden neurons being 100.


For our Deep Neural Network methods we used Deep Neural Network without pretraining (DNN-W), DNN with pretraining using Stacked Autoencoders (DNN-SA) \cite{Hoo_11}, Recurrent Neural Networks (RNN) \cite{Hermans_2013}, RNNs and Long Short Term Memory (RRN-LSTM) \cite{Gers2001}, Convolutional Neural Networks (CNN) \cite{Siripurapu_15} and CNNs and Long Short Term Memory (CNN-LSTM)] \cite{Sainath_15}


To evaluate the goodness of fit of these algorithms we use the Mean Absolute Percentage Error (MAPE) defined as:
\begin{equation} 
\text{MAPE} = \frac{100}{n} \sum_{t=1}^{n} \frac{|y_t - \widehat{y}_t|}{y_t}
\end{equation}
where $n$ is the number of data points, $t$ is the particular time step, $y_t$ is the target or actual value and $\widehat{y}_t$ is the predicted value.

In order to determine the cost of the prediction errors (i.e. whether the prediction is above or below the actual value) the Mean Percentage Error (MPE) is used, which is defined as:
\begin{equation} 
\text{MPE} = \frac{100}{n} \sum_{t=1}^{n} \frac{y_t - \widehat{y}_t}{y_t}
\end{equation}

\subsection{Numerical Results}

\begin{table}[!t]
\caption{DNN algorithms (subscript denotes number of layers)}
\label{tab:2}
\centering
\footnotesize\setlength{\tabcolsep}{3pt}
\begin{tabular}{rrrrrrr}
    \toprule
    Algorithm & \multicolumn{3}{c}{200 Epocs} &\multicolumn{3}{c}{400 Epocs}\\
    \cmidrule(l){2-7}
    & \text{MAPE} & \text{MPE} & \text{Time(s)}  & \text{MAPE} & \text{MPE} & \text{Time(s)}\\
    \midrule
    MLP					& 5.62 	& -5.62 & 14	& 4.55	& -4.54 & 25\\
	$\text{DNN-W}_3$	& \textbf{2.64}	&\text{ }1.61 & 30	& \text{ }2.50 & 1.98	& 56\\
    $\text{DNN-W}_4$	& 5.71	& -5.36 & 37	& 5.48	& -5.32 & 72\\
    $\text{DNN-W}_5$	& 4.40	& \text{ }1.79 & 38	& 5.98	& 5.45 & 69\\
    $\text{DNN-SA}_3$	& 2.97	& \text{ }1.23 & 23	& 2.01	& \text{ }0.74 & 25\\
    $\text{DNN-SA}_4$	& 2.88	& \text{ }0.23 & 29	& 2.37	& \text{ }0.79 & 42\\
    $\text{DNN-SA}_5$	& 2.92	& \text{ }0.91 & 37	& \textbf{1.84} 	& \text{ }0.53 & 49\\
    RNN					& 5.23	& \text{ }0.89 & 174	& 5.13	& -0.37 & 359\\
    RNN-LSTM			& 5.36	& -1.26 & 880	& 5.27	& -1.17 & 1528\\
    CNN-LSTM			& 5.74	& -3.85 & 1029	& 6.43	&  -5.96 & 1912\\
    CNN					& 3.15	& -3.53 & 799	& 4.60	& \text{ }4.23 & 1188\\
    \bottomrule
\end{tabular}
\end{table}

We first look at the baseline methods, (with the exception of MLP) in Table \ref{tab:1}. From the table we see that MLR performs the worst, with a MAPE of 24.25\%, which would indicate that the problem is not linear (see Figure \ref{fig:week_and_weekend}). However, the RT algorithm outperforms the rest of the methods by a noticeable margin. This shows that the problem can be split into some discrete segments which would accurately forecast the load. This can be confirmed by looking at the load  in Figure \ref{fig:week_and_weekend} where it is clear that, depending on the time of day, there is significant overlap of the value of the load between days. Thus, having a node in the RT determining the time of the day would significantly improve accuracy. The run-time for these algorithms was quite short with WMA taking the longest due to the cross-validation step where we determined all possible coefficients in steps of 0.05.

Due to the typically long running time of DNN architectures, the algorithms were restricted to 200 and 400 epocs. From Table \ref{tab:2}, there is a clear difference when looking at the 200 epocs and the 400 epocs MAPE columns, as most of the algorithms have a lower MAPE after running for 400 epocs when compared with 200 epocs. This is especially true for the $\text{DNN-SA}_3$ which saw significant drops in the MAPE. The MLP did not perform the worst in both epocs but it was always in the lower half of accuracy. This indicates that the shallow network might not be finding the patterns or structure of the data as quickly as the DNN architectures. However, it outperformed RT in both the 200 and 400 epocs. This alludes to the fact that the hidden layer is helping to capture some of the underlying dynamics that a RT cannot.



\begin{table}[!t]
\caption{Daily MAPE Values}
\label{tab:3}
\centering
\footnotesize\setlength{\tabcolsep}{3pt}
\begin{tabular}{rrrrrrrr}
    \toprule
    Algorithm & Sun & Mon & Tue & Wed & Thu & Fri& Sat\\
    \midrule
    WMA 	& 5.71	& 10.05 & 8.87 & 10.24 & 10.74 & 10.37 & 10.67\\
    MLR & 65.46	& 27.61 & 12.55 & 11.39 & 9.01 & 9.38 & 35.59\\ 
    MQR & 1.17	& 11.92 & 9.88 & 14.24 & 14.11 & 17.11 & 13.24\\
    RT 	& 7.45	& 5.99 & 7.63 & 7.37 & 5.98 & 7.26 & 8.87\\
    SVR & 20.70	& 12.96 & 10.73 & 11.53 & 11.63 & 10.90 & 17.40\\
    MLP					& 5.18	& 4.62 & 4.43 & 4.27 & 4.31 & 4.70 & 4.34\\
	$\text{DNN-W}_3$	& 2.95	& 1.88 & 2.12 & 2.49 & 2.54 & 2.46 & 3.12\\
    $\text{DNN-W}_4$	& 6.67	& 5.45 & 5.25 & 4.88& 4.61 & 5.65 &5.83\\
    $\text{DNN-W}_5$	& 7.23	&5.53 & 5.56 & 6.14 & 6.13 & 5.81 & 5.48\\
    $\text{DNN-SA}_3$	& 2.29	& 1.84 & 1.76 &1.97& 1.87 & 2.03 & 2.35\\
    $\text{DNN-SA}_4$	& 2.67	& 2.19 & 2.00 & 2.14 & 2.27 & 2.55 & 2.82\\
    $\text{DNN-SA}_5$	& 2.28	& 1.47 & 1.63 & 1.93 & 1.60 & 1.76 & 2.22\\
    RNN					& 5.38	& 5.30 & 4.41 & 5.14 & 5.11 & 5.35 & 5.45\\
    RNN-LSTM			& 4.25	& 4.34 & 4.96 & 4.55 & 5.64 & 6.97 & 6.13\\
    CNN-LSTM			& 7.79	& 6.86 & 6.04 & 6.05 & 5.65 & 6.44 & 6.21\\
    CNN					& 6.39	& 4.20 & 4.27 & 3.32 & 3.87 & 4.18 & 5.03\\
    \bottomrule
\end{tabular}
\end{table}

Looking at the 200 epocs column, we see that $\text{DNN-W}_3$ performs the best with a  MAPE of 2.64\%. On the other hand, the most stable architecture is the DNN-SA with a MAPE consistently less than 3\%. This robustness is shown when the epocs are increased to 400 where the DNN-SA architecture outperforms all the other methods (both the baseline and deep methods). The pretraining certainly gave these methods a boost over the other methods as it guides the learning towards basins of attraction of minima that support better generalization from the training data set \cite{Erhan_2010}. RNNs, and to an extent LSTM, have an internal state which gives it the ability to exhibit dynamic temporal behavior. However, they require a much longer time to compute which is evident in Table \ref{tab:2} since these methods had  trouble mapping those underlying dynamics of the data in such a small number of epocs. CNNs do not maintain internal state, however with load forecasting data, one can expect a fair amount of auto-correlation that requires memory. This could explain their somewhat low but unstable MAPE for 200 and 400 epocs.

Taking both tables into consideration, most of the DNN architectures vastly outperform the traditional approaches, but DNNs require significantly more time to run and thus there is a trade-off. For STLF, which is a very dynamic environment, one cannot wait for a new model to complete its training stage. Hence, this is another reason we limited the number of epocs to 200 and 400. Table \ref{tab:2} shows that limiting the epocs did not adversely affect many of the DNN architectures as most were able to surpass the accuracy of the traditional methods (some by a lot). When selecting a model, one would have to determine if the length of time to run the model is worth the trade-off between accuracy and runtime.

\subsection{Daily Analysis}

\begin{figure}[!t]
\subfloat[Weekday Electrical Usage]{\includegraphics[width=\columnwidth]{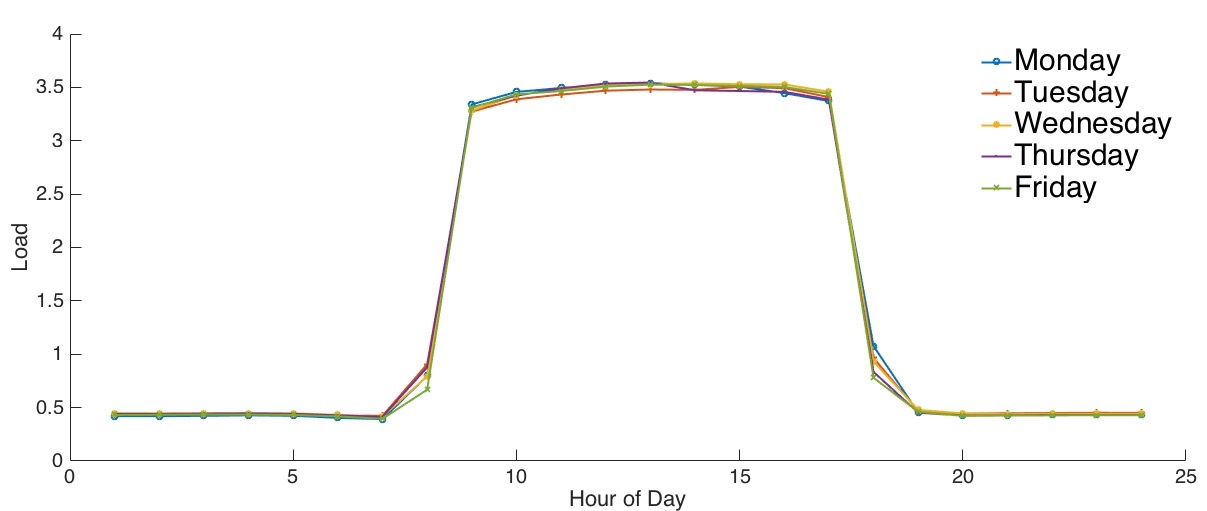}
\label{fig:week}}
\\
\subfloat[Weekend Electrical Usage]{\includegraphics[width=\columnwidth]{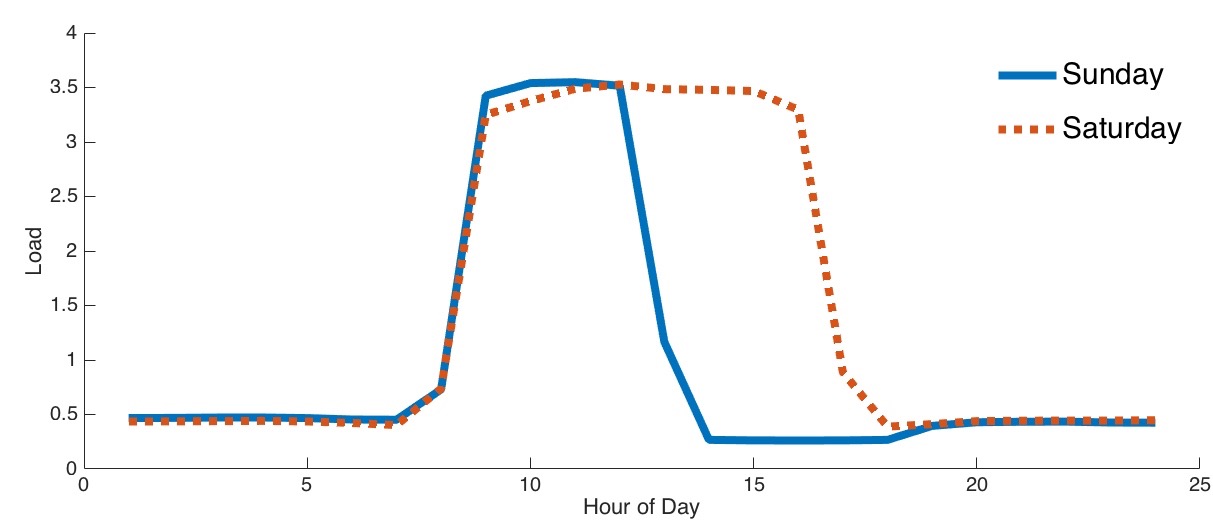}
\label{fig:weekend}}
\caption{Electrical Profiles}
\label{fig:week_and_weekend}
\end{figure}

We know that people have different electrical usage patterns on weekdays when compared to weekends. This difference can be seen in Figure \ref{fig:week_and_weekend} which illustrates usage for a sample home. This household uses more energy during the weekdays than on weekends.  There are electrical profiles that may be opposite, i.e., where the weekend electrical load is more. Whatever the scenario, there are usually different profiles for weekdays and weekends.

To see how our models handle weekdays and weekends, we calculated the average MAPE for each day of the week in the test set (the 400 epoc models was used for the DNNs calculations). The average for each day of the week is tabulated in Table \ref{tab:3}. From the table, it is clear that most of the DNN algorithms have their lowest MAPE during the week. This is indicative that the patterns for weekdays are similar and as a result have more data. By having more data, DNNs are better able to capture the underlying structure of the data and thus are able to predict the electrical load with greater accuracy. Weekend predictions have a higher MAPE since DNNs require a lot of data to perform accurate predictions and for weekends this data is limited. The WMA and MQR seem to have their best day on Sunday, but have a very poor MAPE for the rest of the days. This indicates that the models have an internal bias towards Sunday and as a result fail to accurately predict the values for other days. It is clear, again, that DNNs outperform the traditional methods.

\subsection{Mean Percentage Error}
In this particular domain, an electricity provider will also be interested in changes of electrical load, as opposed to absolute error, in order to adjust generation accordingly, mostly because starting up additional plants takes time. This is why the Mean Percentage Error (MPE) was used. The MPE would tell that a model with a positive value "under-predicts" the load while a negative value "over-predicts" the actual value and they can then adjust their operations accordingly.

Many of the traditional methods had predicted more electrical load than the actual load, including MLP. However, most of the DNNs have under-predicted the load value. Looking at the best in Table \ref{tab:2}, DNN-SAs MPE values (for 400 epocs), they are all under 1\% and positive, which indicates that it under-predicts the value. However, one should not use the MPE alone. An example is RNNs which have a low positive MPE, however it's MAPE in both epocs is around 5\%. This indicates that RNN had a slightly larger sum of values that "under-predicts" than "over-predicts", but its overall accuracy is not as good as other deep architectures.

\subsection{Applications to Energy Efficiency}
Using the results from STLF (MAPE and MPE), a company can now accurately predict upcoming load. This would mean that a power generating company can now produce energy at a much more precise amount rather than producing excess energy that would be wasted. Since most of these companies use fossil fuels which are non-renewable sources of energy, we would be conserving them as well as reducing levels of carbon dioxide released into the atmosphere and the toxic byproducts of fossil fuels.

Another benefit of accurate load forecasting is that of dynamic pricing. Many residential customers pay a fixed rate per kilowatt. Dynamic pricing is an approach that allows the cost of electricity to be based on how expensive this electricity is to produce at a given time. The production cost is based on many factors, which in this paper, is characterized by the algorithms for STLF. By having a precise forecast of electrical load, companies now have the ability to determine trends, especially at peak times. 

An example of this would be in the summer months when many people may want to turn on their air conditioners and thus electricity now becomes expensive to produce as the company could have to start up additional power generating plants to account for this load. If the algorithms predict that there would be this increase in electrical load around the summer months, this would be reflected in the higher price that consumers would need to pay. As a result, most people would not want to keep their air conditioner on all the time (as per usual) but use it only when necessary. Taking this example and adding on washing machines, lights and other appliances, we can see the immense decrease in energy that can be achieved on the consumer side.



\section{Related Work}
The area of short-term load forecasting (STLF) has been studied for many decades but deep learning has only recently seen a surge of research into its applications. Significant research has been focused on Recurrent Neural Networks (RNNs). In the thesis by \cite{mishra_08}, RNNs was used to compare other methods for STLF. These  methods included modifications of MLP by training with algorithms like Particle Swarm Optimization, Genetic Algorithms and Artificial Immune Systems. Two other notable papers that attempt to apply DNN for STLF are \cite{Busseti_12} and \cite{Connor_92}. In \cite{Busseti_12}, they compare Deep Feedfoward Neural Networks, RNNs and kernelized regression. In the paper by \cite{Connor_92} a RNN is used for forecasting loads and the result is compared to a Feedfoward Neural Network. However, a thorough comparison of various DNN architectures is lacking and any applications to dynamic pricing or energy efficiency is absent.
 


\section{Conclusion}
In this paper, we focused on energy wastage in the electrical grid. To achieve this, we first needed to have an accurate algorithm for STLF. With the advent of many deep learning algorithms, we compared the accuracy of a number of deep learning methods and traditional methods. The results indicate that most DNN architectures achieve greater accuracy than traditional methods even when the data is split into weekdays and weekends. However such algorithms have longer runtimes. We also discussed how these algorithms can have a significant impact in conserving energy at both the producer and consumer levels.

\bibliography{data_good}
\bibliographystyle{icml2016}

\end{document}